%% file: main.tex
\documentclass[10pt,twocolumn,letterpaper]{article}

\usepackage[pagenumbers]{iccv} 


\definecolor{iccvblue}{rgb}{0.21,0.49,0.74}
\usepackage[pagebackref,breaklinks,colorlinks,allcolors=iccvblue]{hyperref}

\usepackage{graphicx}

\usepackage{url}

\usepackage{algorithm}
\usepackage{booktabs}
\usepackage{times}
\usepackage{epsfig}
\usepackage{amsmath}
\usepackage{amssymb}
\usepackage[flushleft]{threeparttable}
\usepackage{pifont}

\usepackage[noend]{algpseudocode}

\usepackage{comment}
\usepackage{color}
\usepackage{amsfonts}       
\usepackage{nicefrac}       
\usepackage{microtype}      
\usepackage{xcolor,colortbl}         
\usepackage{tabularx}
\usepackage{epsfig}
\usepackage{lipsum}
\usepackage{pifont}

\usepackage{color}
\usepackage{array}
\usepackage{makecell}

\usepackage[accsupp]{axessibility}  

\def\paperID{4} 
\def\confName{ICCV}
\def\confYear{2025}

\title{OmViD: Omni-supervised active learning for video action detection}

\author{Aayush Rana \qquad Akash Kumar \qquad Vibhav Vineet \qquad Yogesh S. Rawat \\
{\tt\small \{aayushjungbahadur.rana, akash.kumar, yogesh\}@ucf.edu, vibhav.vineet@microsoft.com}
}

\begin{document}
\maketitle

\begin{abstract}
   Video action detection requires dense spatio-temporal annotations which are challenging as well as expensive to obtain.
   However, real-world videos often have varying level of difficulty and may not require equal level of annotations. 
   In this paper we analyze the types of annotation appropriate for each sample and how it affects spatio-temporal video action detection. 
   We focus on two different aspects affecting video action detection; \textbf{1)} how to obtain varying level of annotations for videos, and \textbf{2)} how to learn video action detection with different types of annotations. We study several annotation types including i) \textit{video level tags}, ii) \textit{points} iii) \textit{scribbles}, iv) \textit{bounding box}, and v) \textit{pixel level masks}. 
   First, we propose a simple active learning strategy which estimates appropriate types of annotations required for each video sample.
   Next, we propose a novel learning based spatio-temporal 3D-superpixel approach which generates pseudo-labels from different types of annotations
   and enables learning of video action detection from such annotations. 
   We validate our approach on two different datasets, UCF101-24 and JHMDB-21, for video action detection, significantly reducing the annotation cost without significant drop in performance. 
\end{abstract}

\section{Introduction}

\begin{figure}[h!]
\begin{center}
\includegraphics[width=0.8\linewidth]{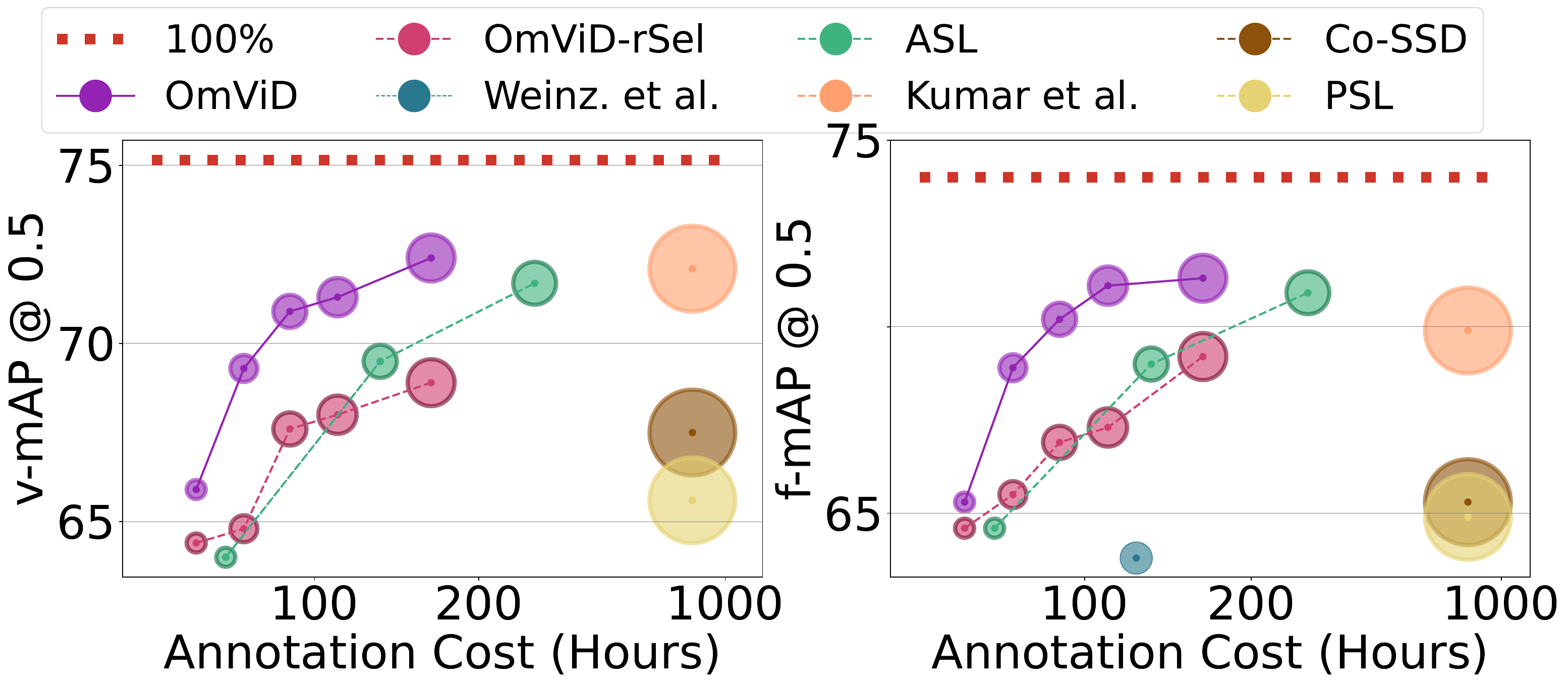}
\end{center}
\caption{OmViD outperforms prior weakly and semi-supervised video action detection methods with lower annotation cost. We report v-mAP/f-mAP @ 0.5 on UCF101 under varying annotation budgets. Bubble size indicates the number of annotated frames per method. Our omni-supervised approach achieves better performance by efficiently annotating more frames at reduced cost.}
\label{fig:teaser_plot_compare}
\end{figure}

\begin{figure*}[!ht]
\begin{center}
\includegraphics[width=0.85\textwidth]{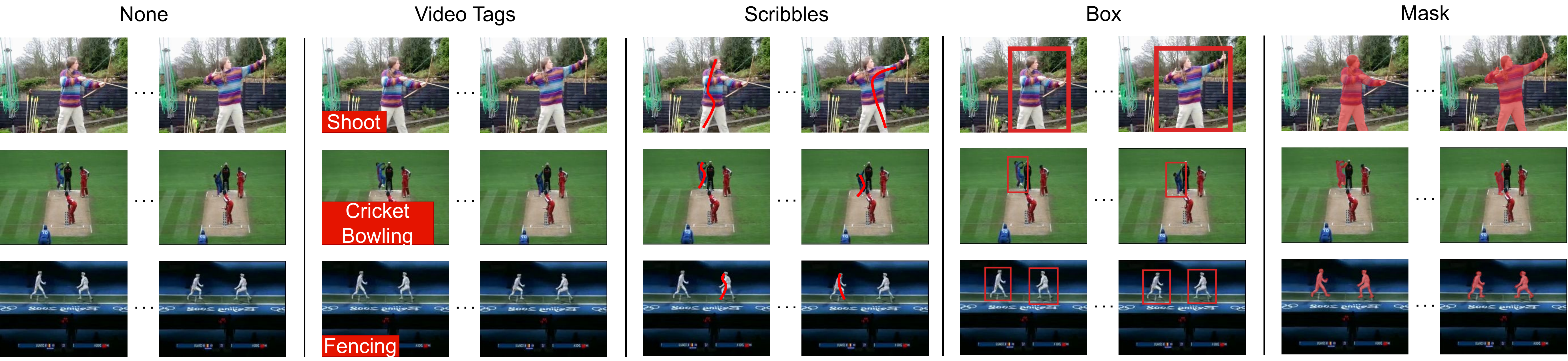}
\end{center}
  \caption{Various types of annotations used for omni-supervised video action detection training. We show two frames from a video sample where each video is annotated with different annotation type.
  }
\label{fig:teaser}
\end{figure*}

Video action detection has been a challenging task as it requires the localization of where an action occurs spatially within a frame as well as when it occurs temporally across multiple frames \cite{hou2017tcnn,soomro2012ucf101,yang2019step,jhuang2013towardsjhmdb,duarte2018videocapsulenet,Dave_2022_WACV,rana2022}. Recent advances in video action detection has stemmed from large models trained on fully annotated data with bounding box and pixel-level masks \cite{yang2019step,zhang2019structured,tang2020asynchronous}. Unlike video action classification task which can be trained with only class labels for the video \cite{zisserman2017quo,hara2018resnet3d,kumar2023benchmarking, Kumar_2025_CVPR}, action detection training requires most of the frames to be correctly annotated. This poses a problem for scaling to large datasets as the number of frames to annotate increases rapidly, limiting big datasets to subsample frames for annotation \cite{gu2017ava,yang2019video}. Lack of large dataset availability also limits the improvement in video action detection task. Therefore, it is important to improve video action detection training using low cost annotation alternatives.

Most recent works using limited annotations for video action detection follow weakly and semi-supervised approach \cite{mettes2019pointly,weinzaepfel2016human,arnab2020uncertainty,cheron2018flexible}. They utilize the ground-truth annotation for a fraction of the training data and generate pseudo-labels for the remaining data. They are often limited by having to rely on external off-the-shelf object detector and post-processing steps (non-maxima suppression, metric based matching) \cite{mettes2017localizing,zhang2019glnet,escorcia2020guess}. These steps require dataset specific parameter tuning for external detector and post processing, further adding complications to the overall training process.
Furthermore, these approaches under-perform significantly compared to supervised training. Recent semi-supervised methods have closed the performance gap \cite{kumar2022end,jeong2019consistency} but still require 20\% of the dataset to have full clean annotations, requiring large annotation cost.

In this work, we study how varying level of annotations 
can be used for video action detection to further reduce the labeling cost.
We focus on two different aspects; 1) how to obtain varying level of annotation for videos, and 2) how to learn video action detection with varying level of supervision from mixed annotation types. We study several levels of annotations; 1) \textit{video level tag}, 2) \textit{point}, 3) \textit{scribble}, 4) \textit{bounding box} and 5) \textit{pixel level mask}. Bounding box and pixel-level mask incurs high annotation cost but provides high quality ground truth which is very effective for model training. Annotations such as scribbles, points and video tags on the other hand induce noise in training due to lack of ground truth information. First, we propose a simple active learning (AL) strategy to estimate the most appropriate annotation for each video which maximizes information gain while using lower annotation cost. Next, we utilize a learnable novel spatio-temporal deep 3D-superpixel module to obtain pseudo-labels. This enables an effective training of action detection models using multiple annotation types at the same time without relying on any external detectors or heuristic based post processing steps.

To summarize, we make the following contributions;
1) We propose a unified \textit{\textbf{Om}ni-Supervised \textbf{Vi}deo Action \textbf{D}etection} method (\textbf{OmViD}) which can utilize different types of annotations 
at the \textbf{same time} to train a video action detection model. 
This approach allows use of \textit{sparse annotations} from varying categories, 
enabling annotation of more videos sparsely instead of having fewer videos with dense per frame annotation for equivalent annotation budget. 2) We propose joint learning of a \textit{\textbf{unified}} pseudo-label estimator together with the action detection for video models, thereby removing reliance on external detector altogether. A novel learning based \textit{\textbf{3D superpixel module}} is used as pseudo-label estimator which predicts spatio-temporal superpixels which are integrated with low cost annotations to estimate the pseudo-labels. 

3) We show that a simple and novel bucket-based active learning strategy for omni-label selection improves performance while keeping the annotation cost low.  
We perform our experiments on UCF101-24 and JHMDB-21 datasets to validate our approach. As shown in Figure \ref{fig:teaser_plot_compare}, we are able to obtain comparable performance with fully-supervised approach using 6\% training data and using fewer annotation cost (\textbf{6\% @ 171 man hours}) compared to prior methods (\textbf{5\% @ 234 man hours, 20\% @ 937 man hours}).


\section{Related work}

\paragraph{Action detection:}
Spatio-temporal action detection in videos is a challenging task where the goal is to detect actions in both space and time for a given video \cite{kalogeiton2017action,peng2016multi,hou2017tcnn}. Prior works utilize the fully supervised action detection approach using dense spatio-temporal annotation (bounding boxes, pixel-wise masks) to train the detection model \cite{yang2019step,li2020actions,duarte2018videocapsulenet}. A common strategy is to use 2D \cite{szegedy2017inceptionv4,he2016deep,simonyan2014very} or 3D \cite{zisserman2017quo,hara2018resnet3d,modi2022video} encoders to get video features and use external region proposals \cite{girshick2015fast,frcnn} for an accurate spatio-temporal action detection in multiple stages \cite{yang2019step,hou2017tcnn}. Alternatively, recent works have shown a single stage anchor-free approach without external proposals also perform competitively \cite{li2020actions,duarte2018videocapsulenet}. Either approach assumes dense annotations present for training, which is a costly in videos for spatio-temporal detection.

\paragraph{Weakly/semi-supervised learning:}
Weakly \cite{khoreva2017simple,singh2017online,arnab2020uncertainty, kumar2025contextual, garg2025stpro} and semi-supervised \cite{Kumar_2022_CVPR,ren2020not,sohn2020fixmatch,berthelot2019mixmatch, Rana_2023_CVPR, kumar2025stable} methods leverage limited annotations via consistency regularization \cite{jeong2019consistency,berthelot2019mixmatch} and pseudo-labeling \cite{xu2021end,li2020improving,rizve2021defense} to train the model at the expense of reduced performance. The annotation cost to performance trade-off is justified as these methods reduce human effort and cost for large scale data annotation for various tasks (classification, localization, detection) \cite{berthelot2019mixmatch,xu2021end,jeong2019consistency,kumar2022end,arnab2020uncertainty,vo2022active}. Recent methods for action detection utilize pseudo-labeling \cite{mettes2017localizing,weinzaepfel2016human}, multi-instance learning \cite{arnab2020uncertainty,mettes2019pointly,mettes2017localizing} and consistency regularization via data augmentation \cite{Kumar_2022_CVPR} to train with limited annotations. However, they often rely on external off-the-shelf object detector \cite{frcnn,Detectron2018,weinzaepfel2016human} or assume availability of a subset of fully annotated data to initialize training \cite{Kumar_2022_CVPR}. 

\paragraph{Omni-supervised learning:}
Most prior work on omni-supervised learning has focused on the image domain, combining various annotation types (e.g., boxes, points, tags, scribbles) to reduce annotation cost \cite{ren2020ufo,wang2022omni,xu2015learning}. Points and scribbles are less expensive to annotate than boxes or masks, yet can still serve as effective pseudo-labels alongside box/mask annotations \cite{mettes2019pointly,ren2020ufo,chen2021points}. Image object detection approaches \cite{yang2019activity,ren2018cross,uijlings2018revisiting,ren2020ufo,wang2022omni} address mixed labels through iterative \cite{uijlings2018revisiting,ren2018cross} or unified frameworks \cite{ren2020ufo,wang2022omni}, often relying on off-the-shelf detectors \cite{yolo9000,frcnn} or pre-computed region proposals \cite{ren2020ufo} for training.

In videos, prior omni-supervised work \cite{cheron2018flexible} uses a single label type per dataset (e.g., boxes, temporal points, tags), supported by external tools like object detectors \cite{Detectron2018}, trackers \cite{lucas1981iterative}, or linkers \cite{kalogeiton2017action,singh2017online}. In contrast, we propose a unified model that leverages mixed annotation types (boxes, masks, scribbles, tags, pseudo-labels) from sparsely labeled data in an end-to-end framework. Our method eliminates the need for external detectors or region proposals by learning a 3D superpixel-based pseudo-labeling approach, extending 2D superpixels from prior work \cite{wang2021ainet,yang2020superpixel}, to support training a robust action detector from heterogeneous supervision.

\paragraph{Active learning:}
Active learning has often been used for budget-aware annotation tasks \cite{bandla2013active,li2013adaptiveAL,joshi2009multi,pardo2021baod,wang2016cost} to obtain training annotations. The general goal of AL is to leverage a scoring mechanism (uncertainty \cite{gal2016dropout,pardo2021baod}, entropy \cite{aghdam2019active,lewis1994heterogeneous}, heuristics \cite{joshi2009multi,kirsch2019batchbald}, set-selection \cite{sener2017active}, clustering \cite{bodo2011active}) which can identify samples most likely to be useful for future training. There has been wide study for image based object classification \cite{li2013adaptiveAL,joshi2009multi} and detection \cite{roy2018deep,choi2021active,yuan2021multiple} to find a new subset of data for annotation, often using an oracle. However, AL for video based task focuses more on classification \cite{vondrick2011video} due to simplicity of the task while only few studies have been done for video action localization \cite{heilbron2018annotate} and action detection \cite{zolfaghari2019temporal}. These AL methods use uncertainty \cite{gal2016dropout,aghdam2019active}, similarity \cite{zolfaghari2019temporal,bodo2011active} or model-estimation \cite{konyushkova2017learning} for sample selection. In this work we extend uncertainty-based method following \cite{gal2016dropout} to videos for sample and frame selection.


\section{Method}

We focus on two different aspects of video action detection: 1) obtaining varying level of annotations, and 2) using these omni-labels effectively for video action detection. We first explain the different types of annotations useful for video action detection. We then explain how we incorporate all the different annotation types into a unified training approach such that the model can learn action detection effectively. Lastly, we explain how to obtain additional data for each type of annotation such that it maximizes performance while keeping annotation cost low.
We estimate cost of each annotation type based on prior works \cite{lin2014coco,wang2022omni,ren2020ufo} (details in supplementary).

\subsection{OMNI-supervision in videos}
Video omni-supervision training assumes presence of multiple types of annotations in the training set, with each sample video having one or more annotation types present at the same time, including tags, points, scribbles, boxes and masks. Unlike images, videos can have multiple frames with sparse annotation of different types. For a given set of training videos $\mathcal{V}$ with $C$ classes, a sample $y_v=\{1...F_v\}$ with $F_v$ total frames can have following annotation types:
\textbf{Video-level tags only:} $y_v={c}$ where $c\in C$ classes. Such samples are only used to train the classifier. 
\textbf{Point annotation:} $y_v={c,p_i}$ where $c\in C$ classes; $p_i=\{P_{x,y}\} \in \{Points\}_{i=1}^{F'}$ are point annotation of $F'$ frames for the sample $y_v$ with $F_v$ total frames.
\textbf{Box annotation:} $y_v={c,b_i}$ where $c\in C$ classes; $b_i=\{x_i^{min},y_i^{min},x_i^{max},y_i^{max}\} \in\{Box\}_{i=1}^{F'}$ are bounding box annotations of $F'$ frames for the sample $y_v$ with $F_v$ total frames. We don't assume the box annotation is present for all frames, so $F'<F_v$ and $|F'|\geq1$.
\textbf{Pixel-wise mask annotation:} $y_v={c,m_i}$ where $c\in C$ classes; $m_i=\{P_{x,y}\} \in \{Mask\}_{i=1}^{F'}$ are pixel-level mask annotation for pixels $P$ of $F'$ frames of sample $y_v$ with $F_v$ total frames (assuming $F'<F_v$ and $|F'|\geq1$). 
\textbf{Scribble annotation:} $y_v={c,s_i}$ where $c\in C$ classes; $s_i=\{P^s_{x,y}\} \in \{Scrib\}_{i=1}^{F'}$ are pixel-level scribble annotation for pixels $P^s$ of $F'$ frames for the sample $y_v$ with $F_v$ total frames. Compared to pixel-wise mask annotation, the set of pixels $P^s$ is smaller than $P$ in mask annotation as scribbles only have pixels of a thin line drawn as scribble. We assume $F'<F_v$ and $|F'|\geq1$. 

\subsection{OmViD: Video action detection}
We propose \textbf{OmViD} as a simplified and unified approach to handle realistic variations in annotations for video action detection task. 
Given a set of training videos $\mathcal{V}$, we assume it has been separated into unlabeled set $\mathcal{D}_U$ and partially labeled set $\mathcal{D}_{PL}$. We use the partially labeled set $\mathcal{D}_{PL}$ which has $B\%$ box (mask) annotation, $S\%$ scribble annotation and $T\%$ video-level tag only annotations to train a unified model $\mathcal{M}$ such that we get prediction $\hat{y}= \mathcal{M}( v; \theta)$ for a given video $v\in\mathcal{D}_{PL}$; where $\theta$ is the trained model weight and prediction $\hat{y}=\{c,Det\}$ for $c\in C$ classes and spatio-temporal detection $Det$.
For samples $t\in T$ with video-level tag only annotations, we assume there are no spatio-temporal annotations present and we only know the video label. However, for samples with box/mask/scribble annotation we assume that we know the video label as well as have at least one spatio-temporal annotation. Furthermore, we also don't restrict the sample to have only one type of spatio-temporal annotation, i.e. a single sample can have some frames annotated with box and some frames annotated with scribbles. Our experiments show that based on the level of utility for different frames in a sample, it is cost effective to have mixed annotation in a single sample.

\paragraph{Pseudo-label generation:}
Superpixels group regions of a given frame $F$ based on similarity in appearance. This property can be leveraged to extend sparse annotations and fill relevant regions of $F$ to generate pseudo-labels for training. With sparse annotation in $F'$ frames for a video with $F_v$ frames such that $F'<F_v$, we will also have to fill connected regions spatially and temporally. 
For this, we propose using learnable 3D superpixels to identify object regions overlapping the sparse annotation spatio-temporally. We identify the 3D superpixels that overlap with the sparse annotation and use all such regions as pseudo ground truth labels. These overlapping regions represent some portion of the object and can give a refined boundary for training purpose. This formulation can learn the 3D superpixel representations from the training data itself in an unsupervised manner and can be jointly trained with the action detection model.

\subsection{Learning objective}

The proposed \textbf{OmViD} is train with mixed type of annotation at the same time; optimizing loss for classification and detection task. Given a video $v$ from the partially labeled set $\mathcal{D}_{PL}$, we optimize the classifier loss $\mathcal{L}_{Cls}$ for all samples as we assume they at least have video-level tags annotation. As we add different types of spatio-temporal annotations, the training process for detection becomes complex as we have to account for pseudo-labels based on the type of annotation.  
Next, we explain in detail about the different losses we use for such unified training.

\subsubsection{Detection loss}
For a given video $v$ with $F_v$ total frames and $F'$ annotated frames such that $F'<F_v$, we first generate pseudo-labels for neighboring frames without spatio-temporal annotations. Since we assume the samples are not densely annotated, we leverage interpolation to get pseudo-labels and use weighted loss to assign appropriate value for each pseudo-label based on their reliability. We give higher weight for pseudo-labels closer to actual ground truth frame and lower weight for those further away. We generalize the detection loss for each sample as,
\begin{equation}
    \mathcal{L}_{Det}= \sum_{i=1}^{F_v} W_i \mathcal{L}_i
    \label{eq:general_detection_loss}
\end{equation}
where, for a sample $v$ with $F_v$ total frames, $W_i$ is the weight of frame $i$ based on closeness to real ground truth and $\mathcal{L}_i$ is the localization loss for frame $i$. For each spatio-temporal annotation type, we compute the loss independently and aggregate them all together. This simplifies the pseudo-label generation as well for each annotation type, specially if a single sample $v$ has multiple annotation type in different frames. Next, we expand the detection loss for each type of annotation and explain the pseudo-label generation process.

\paragraph{Scribble loss:}
The first variation is scribble annotation, where for a given video $v$ we only have scribble annotation $s_i=\{P^s_{x,y}\} \in \{Scrib\}_{i=1}^{F'}$ for frames $F'$, where $P^s$ are pixels with scribbles. 
We use deep 3D superpixels to get $J$ total superpixels such that $\exists s_i \hspace{2mm} \{SPix_j(s_i)\}_{j=0}^J$
; where each superpixel $SPix_j$ has at least one pixel $s_i$ from the scribble in it. This extends the scribble to spatio-temporally connected regions to generate pseudo-labels for detection training.
This allows us to use sparse scribbles to compute the detection loss as,
\begin{equation}
    \mathcal{L}^{Scribble}_{Det} = \sum_{i=1}^{F_v} W_i^{Scribble} \mathcal{L}_i^{Scribble} 
    \label{eq:scribble_detection_loss}
\end{equation}
where, $W_i^{Scribble}$ is the weight for frame $i$ based on closest scribble frame and $\mathcal{L}_i^{Scribble}$ is the localization loss of prediction for $i^{th}$ frame. Next, we describe the superpixel training process which enables pseudo-label generation for scribbles.

\paragraph{3D Superpixel loss:}

Since scribbles are only thin pixels and don't tell us the exact area of the action region in a frame, traditional approaches (interpolation, largest box region) are not able to adequately address the problem of finding right areas to label. This leads us to learn superpixel segmentation to predict connected areas based on their features following \cite{xu2021scribble,lin2016scribblesup}. We propose a learning approach to predict 3D spatio-temporal superpixels and optimize the superpixel loss. First we extend the association map $Q$ for each pixel to predict pixel property $f(p)$, where the property $f(p)$ is a 9-dimensional CIELAB color vector representing a 3D spatio-temporal association map for pixel $p=[x,y,z]$ in a volume. The 3D superpixel loss is given as,
\begin{equation}
    \mathcal{L}_{SLIC}(Q) = \sum_{p} \| f_{col}(p) - f'_{col}(p) \|_2 + \frac{m}{S} \| p-p' \|_2 
    \label{eq:superpixel_loss}
\end{equation}
where, $f_{col}(p)$ is the 9-dimensional pixel property of pixel $p$, $f'_{col}(p)$ is the predicted property for pixel $p$, $S$ is superpixel sampling interval, $m$ is a weight balance, $p$ is the location of pixel in a volume and $p'$ is the predicted location. The first term encourages grouping of pixels with similar interest while the second term enforces spatial compactness of superpixels.

\paragraph{Point supervision:}
In point supervision, for a given video $v$ we only have point annotations for frames $F'$. We follow the 3D superpixel strategy as scribbles to generate pseudo labels and can compute point loss as,
\begin{equation}
    \mathcal{L}^{Point}_{Det} = \sum_{i=1}^{F_v} W_i^{Point} \mathcal{L}_i^{Point} 
    \label{eq:point_detection_loss}
\end{equation}
where, $W_i^{Point}$ is the weight for frame $i$ based on closest point frame and $\mathcal{L}_i^{Point}$ is the localization loss of prediction for $i^{th}$ frame

\paragraph{Bounding box loss:}
For a sample $v$ with box annotations $b_i=\{x_i^{min},y_i^{min},x_i^{max},y_i^{max}\}$ $\in \{Box\}_{i=1}^{F'}$ for $F'$ frames, we generate pseudo-labels for unannotated frames $F''$ such that $F_v=F'\cup F''$. As fine pixel-level boundary is not needed, we use linear interpolation of boxes to get pseudo-labels. The box detection loss is given as,
\begin{equation}
    \mathcal{L}^{Box}_{Det} = \sum_{i=1}^{F_v} W_i^{Box} \mathcal{L}_i^{Box} 
    \label{eq:box_detection_loss}
\end{equation}
where, $W_i^{Box}$ is the weight for frame $i$ based on distance from the closest real box annotation in $\{Box\}_{i=1}^{F'}$, and $\mathcal{L}_i^{Box}$ is the localization loss of the bounding box in frame $i$.

\paragraph{Pixel-wise mask loss:}
For sample with pixel-wise mask annotations, we cannot perform linear interpolation for pseudo-label generation of the mask. Instead, we can compute the pseudo-masks prior to training using pixel-level interpolation. We can then compute the mask loss as,
\begin{equation}
    \mathcal{L}^{Pixel}_{Det} = \sum_{i=1}^{F_v} W_i^{Pixel} \mathcal{L}_i^{Pixel} 
    \label{eq:pixel_detection_loss}
\end{equation}
where, $W_i^{Pixel}$ is the weight for frame $i$ based on closest real mask annotation and $\mathcal{L}_i^{Pixel}$ is the localization loss of predicted mask for $i^{th}$ frame.

\subsubsection{Full training objective}
Based on the individual loss for each annotation type present and the superpixel training, our overall training objective can be specified as,
\begin{equation}
    \mathcal{L} = \mathcal{L}_{Cls} + \lambda_B\mathcal{L}^{Box}_{Det} + \lambda_P\mathcal{L}^{Pixel}_{Det} + \lambda_S\mathcal{L}^{Scribble}_{Det} + \lambda_{Po}\mathcal{L}^{Point}_{Det} + \mathcal{L}_{SLIC}
    \label{eq:overall_loss}
\end{equation}
where, $(\lambda_B, \lambda_P, \lambda_S, \lambda_{Po})\in [0,1]$ are weights that enable the specific loss if a given sample has corresponding annotation.

\subsection{Sample selection}

We start model training using an initial partially labeled set $\mathcal{D}_{PL}^1$. Once the model is trained, we want to increase overall videos by adding some high value samples from the unlabeled set $\mathcal{D}_U$ to get $\mathcal{D}_{PL}^2$. However, there is no specification to the type of annotation each new sample should get. We assume that there is a fixed budget for each round of annotation increment, such that we add $B'\%, S'\%, T'\%$ annotation to the original $B\%, S\%, T\%$ for box, scribble and video-level tag only respectively. We select new videos from $\mathcal{D}_U$ to get $\mathcal{D}_{PL}^2$. We also rank the videos in $\mathcal{D}_{PL}^2$ into buckets to select the annotation type for each video and select the frames to annotate. This selection can be done randomly or using a metric based selection approach. We end up with $\mathcal{D}_{PL}^2$ consisting of $B+B'\%, S+S'\%, T+T'\%$ annotation after one round of sample selection. Next, we describe the bucket-based selection strategy in detail.

\subsubsection{Bucket-based selection strategy}
We use AL to select new videos from $\mathcal{D}_U$. We then rank the videos in $\mathcal{D}_{PL}^2$ into buckets representing each annotation type and select top $\%$ of videos for labeling.

\paragraph{Uncertainty sampling:} 
We follow prior AL methods and use uncertainty based sampling \cite{gal2016dropout,pardo2021baod} to score and select new samples and frames for annotation. Using the model's uncertainty in prediction to score frames, we get aggregated video level scores. Since we get spatio-temporal prediction for each frame, we compute frame-wise uncertainty given as,
\begin{equation}
    \mathcal{U}_f = \sum_{i,j=0}^{I,J} U_{i,j}
    \label{eq:uncertainty}
\end{equation}
where, $\mathcal{U}_f$ is the uncertainty for a frame $f$ with $I\times J$ pixels and $U_{i,j}$ is the uncertainty of model prediction for $i,j^{th}$ pixel. We average score of all frames to get the video level scores used for ranking each video.

\paragraph{Annotation type selection:}
Once we have ranked videos $V_s$ and put in each annotation-type bucket, we combine them with prior partially labeled set $\mathcal{D}_{PL}^1$ to get $\mathcal{D}_{PL}^{2'}$. However, we don't need to fully annotate the new samples from $V_s$, or even give them the same type of annotation. Based on the bucket they fall into after ranking, we simply take top-b\% videos from 'box annotation' bucket and add $b\%$ box/mask annotations to those samples. Similarly we take top-s\% videos from 'scribble annotation' bucket and add $s\%$ scribble annotation and finally repeat same with the 'video-level tag' bucket. The top-b\% and top-s\% is fixed based on annotation budget.


\section{Experiments}

\paragraph{\textbf{Dataset and metrics:}}
We evaluate the proposed approach on UCF-101 \cite{soomro2012ucf101} and J-HMDB \cite{jhuang2013towardsjhmdb} dataset for video action detection. UCF-101 contains 3207 untrimmed videos with spatio-temporal bounding box annotation for 24 action classes. J-HMDB has 928 trimmed videos with pixel-level spatio-temporal annotations for 21 action classes. Following prior action detection work on UCF-101 and J-HMDB \cite{cheron2018flexible,kumar2022end,peng2016multi,weinzaepfel2016human}, we measure frame-mAP and video-mAP scores across different thresholds to measure frame and video level average precision for detections.

\paragraph{\textbf{Training details:}}
We follow \cite{kumar2022end} for the action detection framework, which uses a 2D capsule variant of Video Capsule Network \cite{duarte2018videocapsulenet} with I3D \cite{zisserman2017quo} encoder. This is a simple end-to-end method with single stage encoder-decoder training, which allows us to add the extended 3D superpixel branch from \cite{yang2020superpixel} as an auxiliary decoder. We use the skip-connections from encoder for the superpixel branch directly.
We train the model using a single 16-GB GPU with a batch size of 6 for clips of dimension $8\times 224 \times 224 \times 3$ as $depth \times height \times width \times channels$ as input for the network. 
In our preliminary experiments, we found it very challenging to train an action detection model using point-wise annotation and as such omit it from final model training.
The 3D superpixel branch can be trained in unsupervised fashion. We pretrain the superpixel branch using all frames initially to learn superpixels and use that to generate pseudo-labels for action detection training.

\begin{table*}[t!]
\centering
\caption{\textbf{\textit{Comparison with weakly-supervised methods on UCF-101:}} We show the type of training annotations ($\mathcal{A}$-Type), percent of annotation used as $\mathcal{A}\%$ for available methods and the cost of annotations as ${Cost}\%$. $\mathcal{O}$ denotes if the method uses off-the-shelf object detector to generate training annotations. The annotation types used in training is denoted as: B$\rightarrow$ Box, P$\rightarrow$ Points, S$\rightarrow$ Scribbles, V$\rightarrow$ Video tag. S-20 use semi-supervised approach with 20\% data. We compare f-mAP and v-mAP at different thresholds. We report \cite{weinzaepfel2016human} with their scores for 2 (1.1\%) and 5 (2.8\%) frames annotated per video. OmViD-rSel indicates OmViD with random selection instead of proposed AL.
}
\label{table:baseline_sota_compare_ucf101}
\begin{tabular}{l c|c |r|c|c c c c| c |c c}
\hline
Method & $\mathcal{A}$-Type &$\mathcal{A}\%$ & ${Cost}$ & $\mathcal{O}$ & \multicolumn{4}{c|}{Type} &\textbf{f-mAP@} & \multicolumn{2}{c}{\textbf{v-mAP@}} \\
\cline{5-12}
 &  &  &  & & B & P & S & V & 0.5 & 0.2 & 0.5 \\
\hline
Mettes et al. \cite{mettes2017localizing} & MIX & - &  - & \checkmark & \checkmark & \checkmark &  & \checkmark & - &  37.4 & - \\
Esc. et al. \cite{escorcia2020guess} & Box & - &  - & \checkmark & \checkmark &  &  & \checkmark & - &    45.5 & - \\
Zhang et al. \cite{zhang2019glnet} & Box & - &  - & \checkmark & \checkmark &  &  & \checkmark & 30.4 &        45.5 & 17.3 \\
Arnab et al. \cite{arnab2020uncertainty} & Box & - &  - & \checkmark & \checkmark &  &  & \checkmark & - &     61.7 & 35.0 \\
Mettes et al. \cite{mettes2019pointly} & Point & - &  - & \checkmark &  & \checkmark &  & \checkmark & - &       41.8 & - \\
Cheron et al. \cite{cheron2018flexible} & Box & - &  - & \checkmark & \checkmark &  &  & \checkmark & - &      70.6 & 38.6 \\
Weinz. et al. \cite{weinzaepfel2016human} & Box & 1.1\% & 52 & \checkmark & \checkmark &  &  & \checkmark & - & 57.1 & 46.3 \\
Weinz. et al. \cite{weinzaepfel2016human} & Box & 2.8\% & 132 & \checkmark & \checkmark &  &  & \checkmark & 63.8 & 57.3 & 46.9 \\
ASL \cite{rana2022} & Box & 1.0\%  & 47 & \ding{55} & \checkmark &  &  & \checkmark & 64.7 & 95.3 & 63.9 \\
ASL \cite{rana2022} & Box & 5.0\%  & 235 & \ding{55} & \checkmark &  &  & \checkmark & 70.9 & 96.0 & 71.9 \\

PSL \cite{lee2013pseudo} & Box & S-20\% & 938 & \ding{55} & \checkmark &  &  & \checkmark & 64.9 & 93.0 & 65.6 \\
Co-SSD \cite{jeong2019consistency} & Box & S-20\% & 938 & \ding{55} & \checkmark &  &  & \checkmark & 65.3 &  93.7 & 67.5 \\
Kumar et al. \cite{kumar2022end} & Box & S-20\% & 938 & \ding{55} & \checkmark &  &  & \checkmark & 69.9 & 95.7 & 72.1 \\
Kumar et al. \cite{kumar2025stable} & Box & S-10\% & 469 & \ding{55} & \checkmark &  &  & \checkmark &  73.8 & 95.8 & 76.3 \\

\hline
\rowcolor{cyan!25} Ours (OmViD-rSel) & MIX & 1.0\% & 28  & \ding{55}& \checkmark &  & \checkmark & \checkmark & 64.8 & 95.3 & 64.6 \\
\rowcolor{cyan!25} Ours (OmViD-rSel) & MIX & 6.0\% & 172 & \ding{55} & \checkmark &  & \checkmark & \checkmark & 68.2 & 97.3 & 67.8 \\
\hline
\rowcolor{cyan!25} Ours (OmViD) & MIX & 1.0\% & 28  & \ding{55}& \checkmark &  & \checkmark & \checkmark & 65.3 & 96.1 & 65.9 \\
\rowcolor{cyan!25} Ours (OmViD) & MIX & 6.0\% & 172 & \ding{55} & \checkmark &  & \checkmark & \checkmark & 72.4 & 97.8 & 71.2 \\
\hline
\hline
\rowcolor{yellow!25} Fully supervised baseline & & 100\% & 4.6K & \ding{55} & \checkmark &  &  & \checkmark & 75.2 & 98.8 & 74.0 \\
\end{tabular}

\end{table*}

\begin{table*}[t!]
\centering
\caption{\textbf{\textit{Comparison with weakly-supervised methods on JHMDB}:} We report training annotation type ($\mathcal{A}$-Type), annotated frame percentage ($\mathcal{A}\%$), annotation cost (Cost), and use of off-the-shelf detectors ($\mathcal{O}$). Annotation types: M = Mask, P = Points, S = Scribbles, V = Video tag. S-30 denotes semi-supervised training with 30\% data. We compare f-mAP and v-mAP at various thresholds. OmViD-rSel is OmViD with random selection instead of our proposed active learning.}

\label{table:baseline_sota_compare_jhmdb}
\begin{tabular}{l c|r|r|c|c c c c| c |c c}
\hline
Method & $\mathcal{A}$-Type &$\mathcal{A}\%$ & ${Cost}$ & $\mathcal{O}$ & \multicolumn{4}{c|}{Type} &\textbf{f-mAP@} & \multicolumn{2}{c}{\textbf{v-mAP@}} \\
\cline{5-12}
 &  &  &  & & M & P & S & V & 0.5 & 0.2 & 0.5 \\
\hline
Zhang et al. \cite{zhang2019glnet} & Mask & - & - & \checkmark & \checkmark &  &  & \checkmark & 65.9 & 77.3 & 50.8 \\
Weinz. et al. \cite{weinzaepfel2016human} & Mask & 6\% & 30 & \checkmark & \checkmark &  &  & \checkmark & 50.7 & - & 58.5 \\
Weinz. et al. \cite{weinzaepfel2016human} & Mask & 15\% & 74 & \checkmark & \checkmark &  &  & \checkmark & 56.5 & - & 64.0 \\
ASL \cite{rana2022} & Mask & 6.0\%  & 235 & \ding{55} & \checkmark &  &  & \checkmark & 74.19 & 98.0 & 70.8 \\

PSL \cite{lee2013pseudo} & Mask & S-30\% & 146 & \ding{55} & \checkmark &  &  & \checkmark & 57.4 & 90.1 & 57.4 \\
Co-SSD \cite{jeong2019consistency} & Mask & S-30\% & 146 & \ding{55} & \checkmark &  &  & \checkmark & 60.7 &  94.3 & 58.5 \\
Kumar et al. \cite{kumar2022end} & Mask & S-30\% & 146 & \ding{55} & \checkmark &  &  & \checkmark & 64.4 & 95.4 & 63.5 \\
Kumar et al. \cite{kumar2025stable} & Mask & S-20\% & 97 & \ding{55} & \checkmark &  &  & \checkmark & 69.8 & 98.8& 70.7 \\

\hline
\rowcolor{cyan!25} Ours (OmViD-rSel) & MIX & 6\% & 16 & \ding{55} & \checkmark &  & \checkmark & \checkmark & 56.0 & 97.1 & 57.9 \\
\rowcolor{cyan!25} Ours (OmViD-rSel) & MIX & 15\% & 40 & \ding{55} & \checkmark &  & \checkmark & \checkmark & 67.4 & 97.3 & 67.3 \\
\hline
\rowcolor{cyan!25} Ours (OmViD) & MIX & 6\% & 16 & \ding{55} & \checkmark &  & \checkmark & \checkmark & 58.2 & 97.2 & 56.7 \\
\rowcolor{cyan!25} Ours (OmViD) & MIX & 15\% & 40 & \ding{55} & \checkmark &  & \checkmark & \checkmark & 70.9 & 97.6 & 69.9\\
\hline
\hline
\rowcolor{yellow!25} Fully supervised baseline & & 100\% & 487 & \ding{55} & \checkmark &  &  & \checkmark & 75.8 & 98.9 & 74.9 \\
\end{tabular}

\end{table*}

\subsection{Results}
We evaluate the performance of each dataset trained using various types of annotations mixed at different ratio. We first compare the performance of proposed \textbf{OmViD} method with prior weakly supervised action detection methods and demonstrate the performance for each method along with their annotation properties. Then we evaluate our method at varying levels of supervision to demonstrate the improvement in performance at different level of supervision.

\paragraph{Comparison with prior work:}
We show the performance of our approach in Table \ref{table:baseline_sota_compare_ucf101} and \ref{table:baseline_sota_compare_jhmdb} for UCF-101 and J-HMDB respectively. 
We first compare our approach (OmViD) with prior work on UCF-101 and J-HMDB dataset. Compared to most prior works, we use pseudo-labels from interpolation and 3D superpixels instead of external detector. 
This allows us to use mixed types of annotations, keeping the annotation cost lower than prior approaches for same amount of annotated frames. 
With our omni-supervised training, our total annotation cost for 6\% data (172 hours) is lower than 5\% of prior work (235 hours) while performing better in Table \ref{table:baseline_sota_compare_ucf101}. At 1\% annotations we need only 28 man hours of budget to outperform prior works which need 47 man hours \cite{rana2022} and 52 man hours \cite{weinzaepfel2016human}. Even though \cite{rana2022} provides competitive performance with less annotations, their annotation cost is higher as they do no use omni-supervision, which significantly reduces the annotation cost.
Next, we further analyze the significance of omni-supervision when used with random selection (OmViD-rSel). We observe that even OmViD-rSel also provides competitive performance with minimal annotation cost when compared with existing weakly-supervised methods.

\subsection{Ablations}
\paragraph{Without bucket-based selection:}

\begin{figure}[t!]
\begin{center}
\includegraphics[width=0.95\linewidth]{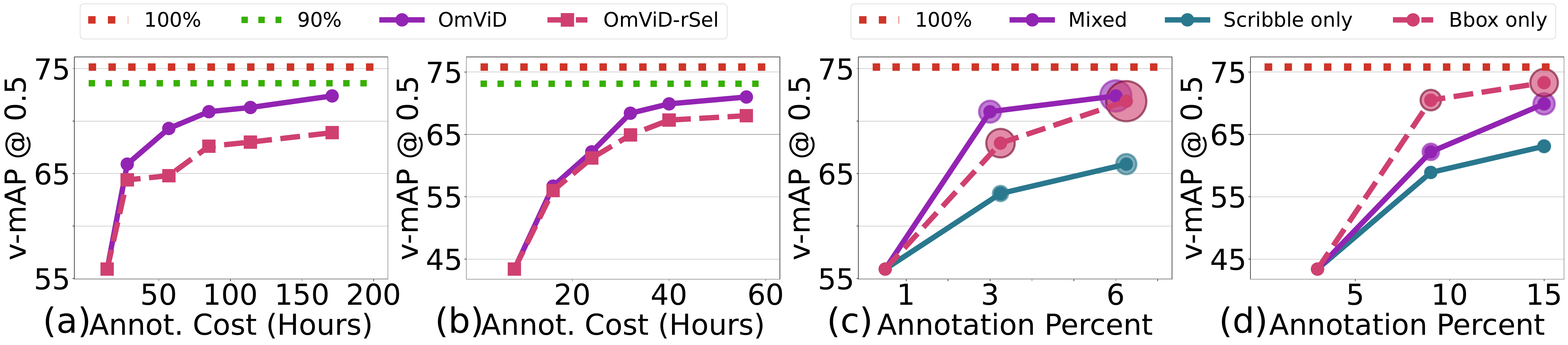}
\end{center}
\caption{\textbf{(a-b)} Comparison of our bucket-based active learning (OmViD) with random selection (OmViD-rSel) on UCF101 (a) and JHMDB (b), highlighting the effectiveness of our frame selection strategy. \textbf{(c-d)} Impact of increasing individual annotation types on performance and cost. Bubble size represents annotation hours per type, showing that increasing only bounding boxes incurs higher cost than scribbles for the same annotation percentage on UCF101 (c) and JHMDB (d).}
\label{fig:ablation_random_vs_AL}
\end{figure}

We evaluate the effectiveness of using proposed bucket-based AL for increasing training annotations. We use random selection for same amount of annotation in each round for fair comparison, so while the annotation cost stays same the frames selected are at random.  
We use an initial set of sparse annotations (bounding-box, mask, tags) equally divided among the videos and increase annotations gradually using proposed bucket-based and random method.
We show the model performance in Figure \ref{fig:ablation_random_vs_AL}(a-b). While performance improves for both selection with more rounds, the model performance on random selection does not improve as much as proposed bucket-based selection for same annotation budget.

\paragraph{Without 3D Superpixel:}
We compare the efficacy of using proposed 3D superpixel to generate pseudo-labels from scribbles with alternative methods for pseudo-label generation. We create pseudo-labels from scribbles by converting it directly to bounding box based on the scribble size as an alternative method. We show the effect of using this alternative pseudo-lable technique on Table \ref{table:ablation_random_bbox_instead_of_scribble} for J-HMDB dataset.
We notice that the bounding box pseudo-labels start falling behind with more annotations added compared to 3D superpixel approach for same annotations. This is primarily due to 3D superpixel generating better annotations from scribbles that is closer to the ground truth.

\begin{table}[t!]
\centering
\caption{We compare OmViD training with and without proposed 3D superpixel on J-HMDB dataset. We fit boxes over scribbles instead of proposed 3D superpixel for pseudo-labels in `wo/Superpixel' method. We report [v-mAP, f-mAP] @ 0.5 IoU scores. 
}
\label{table:ablation_random_bbox_instead_of_scribble}
\resizebox{\linewidth}{!}{
\begin{tabularx}{\linewidth}{@{\extracolsep{\fill}}c|c|c|c c|c c}
\hline
\textbf{Videos} & \textbf{Mask} & \textbf{Scribbles} & \multicolumn{2}{c|}{\textbf{w/ Superpixel}} & \multicolumn{2}{c}{\textbf{wo/Superpixel}} \\
$\%$  & $\%$ & $\%$ & \textbf{v-mAP} & \textbf{f-mAP} & \textbf{v-mAP} & \textbf{f-mAP} \\ 
\hline
30\%    &  1.5\%   &   1.5\%   &   43.4    &   47.1  &   41.8  &  46.6 \\
50\%    &  3.0\%   &   3.0\%   &   56.0    &   57.9  &   58.6  &  61.5 \\
70\%    &  4.5\%   &   4.5\%   &   61.2    &   65.4  &   58.9  &  61.9 \\
90\%    &  6.0\%   &   6.0\%   &   64.9    &   67.3  &   64.2  &  65.4 \\
100\%   &  7.5\%   &   7.5\%   &   67.3    &   67.4  &   65.7  &  66.2 \\
100\%   &  9.0\%   &   9.0\%   &   68.0    &   68.3  &   66.6  &  67.0 \\
\hline 
\end{tabularx}
}
\end{table}

\begin{figure}[t!]
\begin{center}
\includegraphics[width=0.85\linewidth]{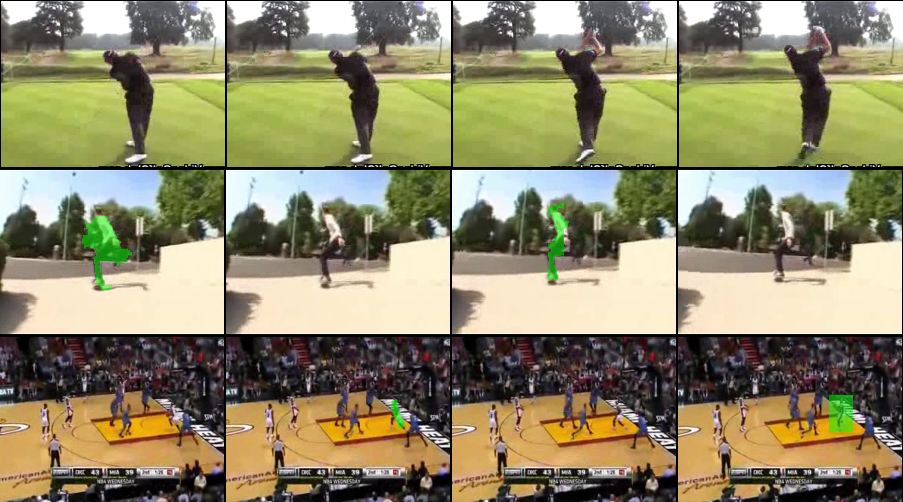}
\end{center}
  \caption{Samples with varying annotation types from our method. 1st row: As GolfSwing is easy class, it only needs video level tag to train this sample. 2nd row: Skating is medium hard class, using more scribble annotations. 3rd row: BasketballDunk is very hard class and uses both box and scribble annotation. 
  }
\label{fig:annot_types_selection}
\vspace{-10pt}
\end{figure}

\paragraph{Annotation type selection:}

We evaluate the importance of having different types of annotations for each round of increment. We use similar setting as baseline but only add scribbles or box/masks in each round instead of proposed mixed increment approach and show the results in Figure \ref{fig:ablation_random_vs_AL}(c-d) for UCF-101 and J-HMDB. We notice that while adding only scribbles has lower cost compared to annotating box/masks, the performance gain is also limited. On the other hand, adding box/mask for all those selected frames in a round would increase performance while also increasing annotation cost. Our proposed approach selects mixed annotation types (box/mask, scribbles, tag only) which reduces annotation budget while still improving performance in a comparable way and provides a quantitative method to trade-off between performance and annotation cost.

\subsection{Analysis and discussion}
\paragraph{Annotation cost reduction:}
Video understanding tasks generally require large annotation budget since each video requires multiple frames annotated. The total annotation budget for UCF-101 with boxes is \textbf{4,686} man hours for 2284 training videos from table \ref{table:baseline_sota_compare_ucf101} and for J-HMDB with masks is \textbf{487} man hours for 666 training videos from table \ref{table:baseline_sota_compare_jhmdb}. This high cost limits dataset scaling compared to image and language domains. 
Our approach uses 171 man hours (vs 235 hours \cite{rana2022}, 938 hours \cite{kumar2022end}) for UCF-101 and 40 man hours (vs 74 hours \cite{weinzaepfel2016human}, 146 hours \cite{kumar2022end}) for J-HMDB and beats prior weakly-supervised work that use more budget for fewer annotated frames.

\begin{figure}[t!]
\begin{center}
\includegraphics[width=0.90\linewidth]{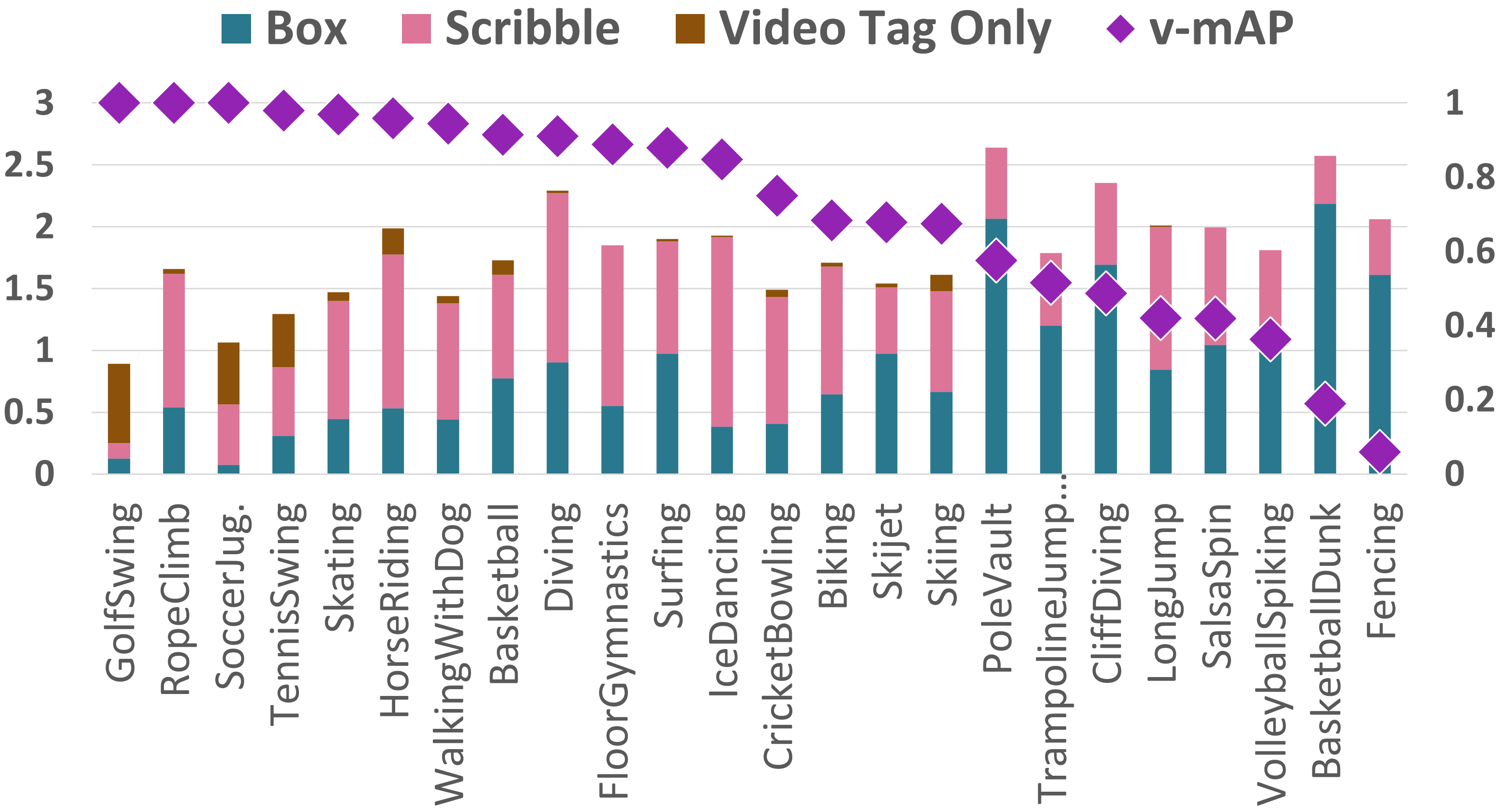}
\vspace{-10pt}
\end{center}
  \caption{Per class annotation distribution for UCF-101. We show the top and bottom 5 performing classes and show the total frames for each type of annotation. The frame count is shown in left axis in thousands (K) and the v-mAP @ 0.5 is shown in right axis. We scale the 'video tags only' by 10 to make it more visible. 
  }
\label{fig:per_class_distribution}
\vspace{-10pt}
\end{figure}

\begin{figure}[t!]
\begin{center}
\includegraphics[width=0.85\linewidth]{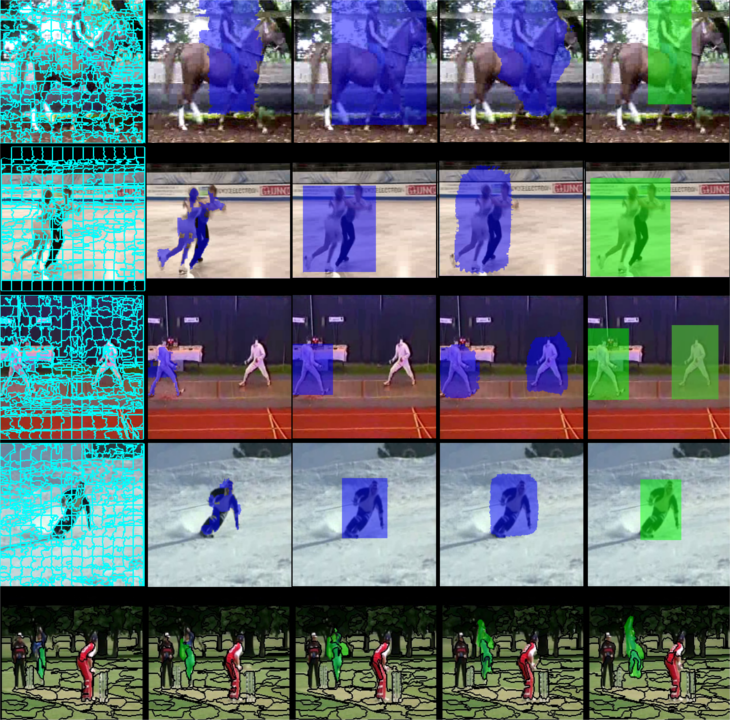}
\vspace{-20pt}
\end{center}
\caption{Illustration of our superpixel-based pseudo-label generation on UCF101. Superpixels (1st column) are aligned with scribbles (2nd column) to generate pseudo-bounding boxes (3rd column), used to train the network for action prediction (4th column). Ground truth is shown in the last column. The final row visualizes the full 3D superpixel label with pseudo-labels highlighted in green.}
\label{fig:scrib_samples}
\end{figure}

\paragraph{Selection approach:}
We provide a quantitative method to select annotations based on the training needs (video tags $\rightarrow$ scribbles $\rightarrow$ box/mask) as shown in Figure \ref{fig:annot_types_selection}. Using AL selection, we pick samples based on their uncertainty (high to low) for box/mask to tag annotation.
We show the selected frame distribution using the proposed approach for UCF-101 in Figure \ref{fig:per_class_distribution}. We notice that well performing classes have fewer box annotation, while low scoring classes have more box annotation. We also notice more scribbles and video only tags for easy classes such as SoccerJug., Skating and GolfSwing. We demonstrate that bucket-based AL selection is a more meaningful use of the annotation budget as we can extract more out of the same annotation budget compared to random selection. Qualitative visuals of proposed approach are shown in Figure \ref{fig:scrib_samples}. 
\vspace{-15pt}

\paragraph{Limitation:}
Unlike prior works that use pretrained detectors with point supervision \cite{wang2022omni,weinzaepfel2016human}, our approach does not rely on such detectors, making point annotations less effective for supervision. Additionally, our 3D superpixel-based pseudo-labeling is challenged by sparse datasets like AVA \cite{gu2017ava}, where frames are annotated only every second.
\vspace{-5pt}


\section{Conclusion}
We present a unified video action detection model that can be trained with different annotation type (box/mask, scribble, tag) under omni-supervision paradigm. It consists of a learnable 3D superpixel module which is jointly trained with the action detection model and helps in generating pseudo-labels from sparse annotations for effective model training. We also demonstrate that proposed model trains better with bucket-based frame selection, utilizing the available annotation budget better by selecting appropriate frames for the respective annotation type.

{
    \small
    \bibliographystyle{ieeenat_fullname}
    \bibliography{main}
}

\end{document}